\documentclass[11pt]{article}

\usepackage[margin=1in]{geometry}
\usepackage[T1]{fontenc}
\usepackage[utf8]{inputenc}
\usepackage{times}
\usepackage{amsmath,amssymb}
\usepackage{booktabs}
\usepackage{array}
\usepackage{enumitem}
\usepackage[hidelinks]{hyperref}

\setlist[itemize]{leftmargin=*, topsep=2pt, itemsep=2pt}
\setlist[enumerate]{leftmargin=*, topsep=2pt, itemsep=2pt}

\newcommand{\code}[1]{\texttt{#1}}

\title{RxScribe Bench: A Multi-Axis Benchmark for Evaluating Vision-Language Models on Indian Outpatient Prescriptions}

\author{
  \href{mailto:somil.dhiman@zstate.ai}{Somil}$^{*}$ \quad
  \href{mailto:vijay@zstate.ai}{Vijay Saini}$^{*}$ \quad
  \href{mailto:vidit@zstate.ai}{Vidit Verma}$^{*}$ \\
  \href{mailto:riya.arora@zstate.ai}{Riya}$^{\dagger}$ \quad
  \href{mailto:aastha@zstate.ai}{Aastha Batta}$^{\dagger}$ \quad
  \href{mailto:vibhuti.malhotra@zstate.ai}{Vibhuti Malhotra}$^{\dagger}$ \\
  \href{mailto:chayan.khetan@zstate.ai}{Chayan Khetan}$^{\ddagger}$ \quad
  \href{mailto:piyush.mittal@zstate.ai}{Piyush Mittal}$^{\ddagger}$ \quad
  \href{mailto:puneet.poonia@zstate.ai}{Puneet Poonia}$^{\ddagger}$ \\[6pt]
  \small Zstate AI
}
\date{}

\begin{document}
\maketitle
\begin{center}
\small $^{*}$Equal contribution.\quad $^{\dagger}$Data operations and quality control.\quad $^{\ddagger}$Engineering infrastructure.
\end{center}

\begin{abstract}
Prescription transcription errors are not interchangeable. A model that fabricates a drug and a model
that misreads a legible dose pose very different clinical risks, yet prescription-transcription accuracy
is typically reported as a single blended figure that treats the two as equivalent. We introduce
\textbf{RxScribe Bench}, a benchmark for evaluating vision-language models on handwritten prescription
digitization that decomposes performance into four axes tied to clinical severity, rather than folding
everything into a single aggregated score. Given only a prescription image and an output schema, a model
produces a structured record, which is then compared field by field against a human-authored ground
truth of identical shape, with each field also labeled for visibility and legibility. The four axes
isolate distinct failure modes, namely \emph{Correctness}, \emph{Hallucination}, \emph{Engagement}, and
\emph{Robustness}. The Robustness axis withholds its hard-field results rather than reporting one when
the supporting sample falls below a minimum-evidence threshold. We evaluate frontier vision-language models on real
prescriptions across independent cold runs per image, and find that no single model wins across all
four axes.
\end{abstract}

\section{Introduction}

Illegible handwriting is a well-documented source of transcription error in the OCR and
handwritten-text-recognition literature \cite{ocr-cer,htr-kvie}, and a prescription is an unusually unforgiving
instance of the problem. A dose read as 500 where the page says 50, or a brand resolved to the wrong
molecule, reaches the pharmacist looking exactly like a correct reading. Reading a prescription is not a single skill. It's a collection of skills: figuring out what’s on the page, writing it down once you've figured it out, interacting with difficult-to-read content rather than skipping it, and guessing at the pharmacology that the page doesn’t tell you. These are separable skills, and a system can fail at one while succeeding at the others. A model that garbles a clinic’s address and a model that invents an active ingredient may be equally accurate by count, but pose different risks in a pharmacy.The question this raises is not how much a model gets right, but what kind of mistake it makes when it is wrong.

Existing evaluation practice does not answer that question directly. General document
visual-question-answering benchmarks \cite{docvqa} score text-extraction accuracy but do not distinguish
a fabricated answer from a declined one, or weight an error by its consequence. Prior benchmarking work
on prescription OCR itself \cite{ocr-cer} reports one error rate, typically character error rate or
word error rate, over the whole document; a misread letterhead and a misread dose cost the same, and a
transcription slip counts identically to an invented word. And recent analyses of multimodal benchmark
design \cite{redundancy1,redundancy2,redundancy3} find that many ostensibly distinct benchmarks in fact
measure the same underlying capabilities, so a set of axes has to earn its separateness rather than
assume it.
Vision-language hallucination benchmarks \cite{amber,hallusionbench} already report multiple dimensions
rather than one blended score, but these dimensions are not tied to clinical severity and are not
evaluated on this domain. None of these tie the type of error to its clinical consequence in this
domain. The one exception that evaluates this domain directly, \cite{mirage}, does so after fine-tuning
rather than zero-shot, which is the setting this paper targets.

RxScribe Bench is built to answer that question directly. We make four contributions.

\begin{itemize}
  \item A schema-aligned evaluation of vision-language models on handwritten prescription digitization,
  in which a model's structured output is compared field by field against a human-authored ground truth
  of identical shape.
  \item Four axes reported separately rather than blended, tied to clinical severity rather than to a
  single notion of accuracy, namely \emph{Correctness}, \emph{Hallucination}, \emph{Engagement}, and
  \emph{Robustness}.
  \item A minimum-evidence reporting threshold on the Robustness axis's hard-field results, under which a
  statistic computed from an insufficient number of observations is withheld rather than reported, since
  it would otherwise appear with the same apparent precision as a statistic supported by substantially
  more evidence.
  \item An evaluation of frontier vision-language models on real prescriptions across independent cold
  runs, showing that no single model wins across all four axes.
\end{itemize}

\section{Task Definition and Data Provenance}

\subsection{Data source, de-identification, and use permission}

The prescription images used in this benchmark were collected individually from outpatient clinics in
India, with the permission of the prescribing clinicians and clinics that produced them, rather than
drawn from an existing institutional dataset. All prescription images were manually reviewed for personally identifiable information (PII) prior to
inclusion in the benchmark. Identifying information, including patient names, contact details,
addresses, identification numbers, and other potentially sensitive information, was manually identified
and anonymized before any annotator or model saw an image, leaving only the clinically relevant content
the task requires: medications, dosages, diagnoses, and the administrative fields. A subsequent manual
review was performed to verify that the de-identified images contained no residual personally
identifiable information.

\subsection{Model input and schema}

A model receives one prescription image and a JSON Schema listing the fields to fill and their types,
and returns a single JSON object conforming to that schema. The schema
carries no ground-truth values, worked examples or hints about the image's contents, and the pipeline
adds no OCR pre-pass, embedded text layer, retrieval step or second attempt after a failed one. Every character
in the output is therefore either read from the image or supplied from the model's own knowledge, and
exactly one field draws on the latter, the active ingredient, rarely printed on a prescription since
only the brand name usually is, which the model must infer pharmacologically rather than read. Every
other field is a transcription task.

The prompt template, the model's output and the human-authored ground truth all follow the same
JSON layout, so any two of them can be overlaid and compared field for field with no mapping step in
between. The layout has three shapes: scalar fields, list sections that hold an arbitrary number of
same-typed entries, and one table, the medicine list, in which each row carries a fixed set of columns:
the drug's name as written, a corrected reading of that name where it needs one, the inferred active
ingredient, frequency, duration, dosage form and relation to meals. Every section is assigned, once, to exactly one of three tiers. \emph{Medication} is the
medicine table. \emph{Clinical} covers everything else that describes the patient or the treatment:
diagnoses, symptoms, vitals, history, allergies and recommended tests. \emph{Non-clinical} covers the
administrative surround: the prescriber's name and registration, the facility and the document date. A
section not named in any tier defaults to non-clinical, so adding a field to the schema can never
silently enlarge the medicine numbers.

\subsection{Ground-truth annotation and quality assurance}

Each image has a single ground-truth file that follows the same schema and records the visibility, legibility, and value of each field. Every field carries exactly one of four legibility labels: Clear, Partial, Blur, or Not Visible. Later sections refer to a field labeled Partial or Blur as a \emph{hard} field, a field labeled Clear as a \emph{clear} field, and a field labeled Not Visible as an \emph{absent} field; these are this paper's own shorthand for the same four labels, not a different classification. Fields marked as Not Visible are still evaluated for presence; therefore, assigning a value to such a field is considered a fabrication. Similarly, for fields marked as Blur or Partial, leaving the field blank is considered a miss rather than a valid abstention, as the annotation indicates that the relevant information is recoverable. Each image was independently annotated by two annotators, and the resulting annotations were compared using percentage exact-match agreement, which reached 96\% across the complete dataset. When the annotators disagree, a reviewer independently assesses the available evidence and resolves the discrepancy. The reviewer remains blinded to which annotator provides each value and bases the decision solely on the supporting evidence notes.

\section{Evaluation Protocol}

Each image is evaluated independently. The model receives only the image and the corresponding schema-derived prompt, with no additional information or context. The request is routed to the appropriate provider based on the specified model identifier. The scoring process is handled separately from the model call to ensure that the evaluation remains independent of the inference process. Ground truth never enters live model
querying, so a model can never leak the answer key through its own run. The scoring routine is a pure function of a prediction and a ground-truth record, with no image,
no network call and no side effects, because prediction
and ground truth share one shape. Both trees are first flattened into aligned scalar and array units; each unit
is compared by type. Those per-field scores are then regrouped into the four published axes,
and per-record axis values are rolled up across runs and images into the per-model numbers reported
below. Nothing is re-inferred or re-scored to build the benchmark. It is a pure regrouping
of records already on disk, verified against those stored per-field scores. A response that fails to
parse into the required JSON schema is not retried and is not dropped from the pool; it is scored like
any other prediction, so a schema violation yields no predicted fields and every ground-truth field on
that record is charged as a miss, treating a parse failure as a capability signal rather than an error
to be papered over.

\section{Field-Level Scoring Rules}

Every axis is built from the same handful of rules, applied to one field at a time. What follows says
what each rule measures and why it works that way, not how the scoring implementation is built field by
field; the exact constants, thresholds and tolerances that implementation uses live with the scoring code
rather than in this text. This is a deliberate scope boundary rather than an omission: Section~3 already
establishes that every number in this paper is a regrouping of records that implementation produced, so
the implementation, not this section, is the artifact a reader needs for bit-exact reproduction.

\subsection{Free-text fields}

A predicted string is compared to the ground-truth string by ANLS, a normalized
edit-distance similarity \cite{anls-impl}, after lower-casing, collapsing whitespace, and folding a
small closed set of glyphs onto one representative each, so that a small misspelling costs a little and
a wrong answer costs nearly everything. The fold is an encoding normalisation, not a semantic one: it
covers the separator strokes, the comparison symbols and the connector an annotator draws between a
label and its value, each of which reaches the file in several code points that carry no difference in
reading. A comparison symbol is mapped onto a single representative rather than removed, so a stated
clinical threshold survives normalisation intact and a prediction that omits the operator is still
charged for omitting it. The connector, carrying no reading of its own, is reduced to a separator so
that the tokens on either side of it neither fuse nor are compared to a stray mark. The fold applies to
prediction and ground truth alike, so neither side is advantaged by which encoding it happens to use. If the model asserts nothing
where the ground truth holds a value, the field earns nothing and is counted a miss.

\subsection{Optimal-assignment matching}

Pairing a predicted item to a ground-truth one is a different
question from grading it, and that pairing runs as an optimal one-to-one assignment across the whole
table or list, rather than first-come matching that an early bad guess could derail. Medicine rows are
paired on a canonicalized form of the drug name that disregards dosage-form and route tokens, so a
predicted row and a ground-truth row differing only by such a token still pair with one another, and
the model is charged neither with inventing a drug nor with missing one; the name itself is still
graded on what the model actually wrote, so the stray token costs marks on the name specifically and
nowhere else. A model that
merges two ground-truth entries into a single string earns partial credit for whichever entry it most
resembles and is charged a miss for the other, which exposes a list-merge as the structural error it is
instead of disguising it as a spelling error.

\subsection{Medicine-cell grading}
\label{sec:medicine-cell-grading}

The clinical scoring used for the published evaluation axes treats safety-relevant outcomes as binary, meaning that a near-miss does not receive partial credit. This distinction is intentional, as partial credit on a reading-related metric should not influence a safety-critical judgment. At the same time, the extent of the error is measured continuously for diagnostic purposes, allowing near-misses to provide a more detailed view of model performance. Generic-name inference is handled separately because it is a knowledge-based task rather than a direct reading task. In most prescriptions, the active ingredient is not explicitly stated, and the medication is usually identified by its brand name. Identifying the corresponding generic name therefore requires pharmacological knowledge beyond the information explicitly stated on the prescription. For this reason, generic names are not evaluated by surface-level string similarity over the whole field, which would penalise differences that carry no clinical meaning. The field is instead compared as an unordered set of ingredients, so the order in which a combination product's components are written does not affect the score.

Equivalence between two names for the same ingredient is declared by the ground truth itself. Each medicine row whose composition is determinate carries, alongside the transcribed generic field, the ingredient list that row is understood to contain and, for each ingredient, the alternative names that row accepts for it. An accepted alternative may name the same substance under another name, spelling or abbreviation, a salt, ester or hydrate that delivers the same active moiety, or a broader parent that subsumes the ingredient; it may not name a separately marketed product in the same class. Because these declarations travel with the annotation rather than living in the scoring code, a record re-scores identically at any later date, and the equivalences a number depends on can be audited alongside the ground truth that produced it. Rows whose composition is indeterminate, being excipient-heavy or otherwise not reducible to a definite ingredient set, are exempt from ingredient scoring entirely rather than scored against a list no reader could reproduce; the exemption is an omission, not a zero, so an exempt row neither earns nor forfeits credit on this field.

\section{The Four Axes}

The benchmark reports performance across four separate axes, without combining them into a single overall score. Each axis is reported as a set of values rather than a single number, allowing the evaluation to preserve important differences in model performance.

\subsection{Correctness}

Correctness measures how much of the information in the prescription the model identifies correctly. Credit requires both that a field was
filled in and that it was read correctly; a miss and a fabrication both cost marks. For a field $k$,
$c(k)$ is the credit that field earns, ranging from none for a wrong or missing value up to full credit
for an exact match. Let $\mathcal{A}$ be the set of fields the model asserted a value for, with
$N_{\mathrm{asserted}} = |\mathcal{A}|$, and let $\mathcal{T}$ be the set of fields the ground truth
actually holds a value for, with $N_{\mathrm{present}} = |\mathcal{T}|$. Precision, $P$, sums credit over
$\mathcal{A}$ and divides by $N_{\mathrm{asserted}}$. Recall, $R$, sums the same credit over
$\mathcal{T}$ and divides by $N_{\mathrm{present}}$. $F$ is the harmonic mean of $P$ and $R$, so it only
climbs when both climb together, and a model cannot buy a high harmonic mean by inflating one at the other's
expense.
\[
P = \frac{\sum_{k \in \mathcal{A}} c(k)}{N_{\mathrm{asserted}}}
\]
\[
R = \frac{\sum_{k \in \mathcal{T}} c(k)}{N_{\mathrm{present}}}
\]
\[
F = \frac{2PR}{P+R}
\]

This $F$ score is calculated once per run. A prescription's score is the unweighted mean of its three
runs' scores, and the single number reported below is in turn the unweighted mean of those
per-prescription scores across every image, so every image counts equally regardless of how many fields
it carries. A second variant is tracked internally, which pools every run's and every image's underlying
credit and denominator sums into one global ratio before computing a single $F$ score, weighting each
image by its field count rather than counting it once; that pooled variant is not the number shown in the
tables below. Precision and recall above are defined to show what the harmonic mean is balancing, not
because both are separately published in the tables: the number reported there is a mean of per-run $F$
scores, which is not itself derived from any single precision/recall pair, so there is no precision or
recall to report alongside it. The pooled variant just described does yield a single global precision and
recall, but, like the pooled $F$ it produces, it is tracked only as an internal diagnostic rather than
published. This score is further split in two: a medicine-only score computed over the medicine table alone,
and a clinical-plus-non-clinical score that pools the clinical tier, meaning diagnoses, symptoms, vitals,
history, allergies and recommended tests, together with the non-clinical tier, meaning the prescriber,
facility and document fields. The benchmark does not currently report the clinical tier on its own.

\subsection{Hallucination}

Hallucination quantifies what the model asserted with no support in the source document. The active
ingredient is a deliberate exception to that definition: it is the one field the model must infer
pharmacologically rather than read (Section~2.2), so the document never supports it either way, and every
value asserted there, correct or not, is technically unsupported by the page. What the tiers below
actually charge on that field is therefore not the act of inferring it but an incorrect inference: an
asserted active ingredient counts as a fabrication only when it fails to match the ingredient list and
the accepted alternatives that row declares (Section~\ref{sec:medicine-cell-grading}), not merely
because the document itself never states it.

One further class of event is excluded by construction. Where ground truth holds a field as a list and
a model returns the same content in a different container shape, the mismatch is a structural error and
is charged as such by the list comparison; it is not additionally counted here, since the model asserted
a field the page does support and the value it asserted is not unsupported. Counting it would charge one
structural mistake twice and would attribute a fabrication to a field the ground truth demonstrably
holds. It
is reported
as counts within three disjoint severity tiers rather than as a single blended rate, since a fabrication
in one tier carries a materially different consequence than a fabrication in another, and a single rate
cannot indicate whether the underlying sample consists of a handful of events or many. Each tier's count
is further normalized by the number of valid runs, so that models with different run totals remain
comparable, and the single blended rate this tiering replaces is retained alongside it for continuity,
explicitly labeled as the superseded metric so the earlier and current views can be compared directly.
The tiers above count fabrication \emph{events}, one per invented entry, whereas the superseded flat rate
instead pools true and false positives at the level of every individual scalar field and every individual
medicine-table column. Its false-positive tally is
therefore a finer-grained count than the Total column below, and a reader cannot recover the flat rate by
dividing Total by Asserted; Asserted is the flat rate's own denominator, but its numerator is a separate
field-level tally not shown elsewhere in the table:

\begin{table}[t]
\centering
\small
\begin{tabular}{@{}l p{5.8cm} p{4.3cm}@{}}
\toprule
Tier & What counts & Why it sits here \\
\midrule
T1, fabricated medications & invented drug row + invented active ingredient on a real row & can directly harm a patient \\
T2, fabricated clinical values & clinical-tier field filled where blank, or invented lab-test entry & clinically misleading \\
T3, fabricated non-clinical fields & invented administrative/prescriber metadata & wrong, not dangerous \\
\bottomrule
\end{tabular}
\caption{Hallucination severity tiers.}
\label{tab:hallucination-tiers}
\end{table}

\[
\text{Fabrications per run} = \frac{\text{Total fabrications}}{\text{Number of valid runs}},
\qquad
\text{Flat rate} = \frac{\text{False positives}}{\text{True positives} + \text{False positives}}
\]

\subsection{Engagement}

Engagement measures whether the model fills in the fields that are actually on the sheet, and stays quiet
only where there is nothing to read. The same trade-off, abstaining rather than answering incorrectly, has
been studied directly for visual question answering \cite{reliable-vqa} and surveyed more broadly across
language models \cite{abstention-survey}. Every hard field, meaning one the annotator labeled Partial or Blur (Section~2.3), contains information that a human annotator is able to recover. Therefore, leaving such a field blank is treated as a miss rather than as a cautious response. The axis reports performance across three populations separately, hard fields, clear fields, and absent fields, since leaving a field blank is beneficial only when the field is genuinely absent.

\subsubsection{1. Hard Fields Present}

$$
\mathrm{Attempt}_{\mathrm{hard\ present}}
=
\frac{\text{number of hard fields attempted}}
{\text{number of hard fields present}}
$$

\subsubsection{2. Clear Fields Present}

$$
\mathrm{Attempt}_{\mathrm{clear\ present}}
=
1 -
\frac{\text{number of clear fields left blank}}
{\text{number of clear fields present}}
$$

\subsubsection{3. Fields Absent}

$$
\mathrm{Silent}_{\mathrm{absent}}
=
\frac{\text{number of absent fields left blank}}
{\text{number of absent fields}}
$$

The primary measure is $\mathrm{Attempt}_{\mathrm{hard\ present}}$, which captures how often the model attempts to recover information from a field that is present but difficult to read, rather than leaving it blank. $\mathrm{Attempt}_{\mathrm{clear\ present}}$ provides the
reference against which that rate must be read, because a low hard-field attempt rate is evidence of
selective caution only if the clear-field attempt rate is high; if both are low, the model is not being
selectively careful but uniformly unresponsive. $\mathrm{Silent}_{\mathrm{absent}}$ isolates the one
population of the three in which withholding a value is the correct behavior, since no content exists
there to report; asserting a value on such a field is instead counted as a fabrication. Each of the three
quantities is computed over its own population of fields. Engagement applies no minimum-evidence floor of
its own: unlike Robustness below, every rate here is reported regardless of how small its underlying
population is, so a rate drawn from a small population should be read with the appropriate caution rather
than expecting it to be withheld automatically.

\subsection{Robustness}

Robustness looks at how the model holds up on messy handwriting. Two things are asked and never
combined: how well it does on the difficult fields in absolute terms, and how much it loses relative to
clean handwriting. Combining them misleads in both directions, because a model that reads everything
badly has a small drop and would look robust, while a strong model with a large drop can still beat it
outright on the hard cases. Each is published twice, because averaging only the fields a model chose to
answer is a selection effect rather than a reading score:
\[
\begin{aligned}
\mathrm{hard\ delivered} &= \frac{\text{credit earned on hard fields}}{\text{hard fields present}} \\[4pt]
\mathrm{hard\ attempted} &= \frac{\text{credit earned on hard fields}}{\text{hard fields answered}} \\[4pt]
\mathrm{drop} &= \mathrm{clear} - \mathrm{hard} \\[4pt]
\mathrm{parity} &= 1 - \bigl| \mathrm{Attempt}_{\mathrm{hard\ present}} - \mathrm{Attempt}_{\mathrm{clear\ present}} \bigr|
\end{aligned}
\]
Delivered is the main version, and it treats a skipped hard field the same as a wrong answer, which is
fair since every hard field had something in it a human could read. Attempted-only only counts the
fields the model actually answered, and we keep it as a secondary check, since it shows whether a model
couldn't read a field rather than simply chose not to try. The attempted-only score can be misleading if a model avoids fields it finds difficult and answers only the easier ones. Engagement parity addresses this by comparing how often the model attempts hard fields with clear fields. A large gap indicates that the attempted-only score may overstate the model’s ability to handle difficult fields. Partial credit is retained for near-misses, while hard-field results are withheld rather than
reported when a model's hard-field population falls below a minimum-evidence threshold of 5 fields; no
model in this evaluation falls anywhere near that threshold, so the withholding rule does not visibly
fire in the tables below, but it remains active for any future run where a model's hard-field sample is
this thin.

\section{Results}

\subsection{Experimental Setup}

This evaluation covers 200 prescription images, four frontier models, and three independent cold runs
for each model on each image, giving 2,400 scored runs and 600 runs per model. Each run is evaluated
against a shared set of ground-truth fields; the pooled ground truth holds 23,913 scored units, of which
12,288 fall in the medicine table and 11,625 outside it. All models receive the same images and schema
and generate their outputs independently. Every number below is produced by the deterministic scorer
alone, with no model-based judge in the loop, so the whole table set is reproducible from the stored
predictions and ground truth without any further inference.

\begin{table}[t]
\centering
\small
\begin{tabular}{@{}l l@{}}
\toprule
Model & Provider \\
\midrule
\code{claude-opus-5} & Anthropic \\
\code{gemini-3.1-pro-preview} & Google \\
\code{muse-spark-1.2} & Meta \\
\code{gpt-5.6} & OpenAI \\
\bottomrule
\end{tabular}
\caption{Models evaluated and their provider.}
\label{tab:models}
\end{table}

\subsection{Correctness}

\begin{table}[t]
\centering
\footnotesize
\setlength{\tabcolsep}{4pt}
\begin{tabular}{@{}l r r r@{}}
\toprule
Model & Correctness & Medicine-only & Clinical + Non-clinical \\
\midrule
GPT-5.6 & 68.10\% & 57.83\% & 77.84\% \\
Gemini 3.1 Pro & 66.97\% & 55.06\% & 78.94\% \\
Muse Spark 1.2 & 64.45\% & 49.49\% & 78.87\% \\
Claude Opus 5 & 61.67\% & 47.08\% & 75.59\% \\
\bottomrule
\end{tabular}
\caption{Correctness axis: pooled $F$ score, overall and split by tier (medicine-only vs.\ clinical
tier and non-clinical tier pooled together). The clinical and non-clinical tiers are not broken out
separately.}
\label{tab:correctness}
\end{table}

GPT-5.6 leads overall correctness and the medicine-only split, the latter by a clear margin over the
next model. Gemini 3.1 Pro leads the pooled clinical-plus-non-clinical split, but only just: Muse
Spark 1.2 sits within a tenth of a point of it there, so that particular lead should not be read as a
separation between the two. No model leads every split, and Claude Opus 5 trails on all three.

Every model reads the medicine table markedly worse than the rest of the sheet, and the ordering of the
four models is not the same in the two splits: the gap between a model's medicine score and its
clinical-plus-non-clinical score ranges from twenty to twenty-nine points and is widest for the two
weakest readers overall.

The medicine split and the pooled clinical-plus-non-clinical split are not directly comparable:
medicine-table cells are graded pass/fail under Section~\ref{sec:medicine-cell-grading}, so a near-miss counts as a complete miss,
while clinical and non-clinical fields are scored by ANLS and can still earn partial credit for a close
reading. The lower medicine score therefore reflects this difference in grading rules at least as much
as any difference in reading accuracy.

\subsection{Hallucination}

\begin{table}[t]
\centering
\footnotesize
\setlength{\tabcolsep}{4pt}
\begin{tabular}{@{}l r r r r r r r@{}}
\toprule
Model & T1 (meds) & T2 (clinical) & T3 (non-clin.) & Total & Per run & Asserted & Flat rate \\
\midrule
GPT-5.6 & 600 & 790 & 381 & 1{,}771 & 2.95 & 16{,}722 & 8.91\% \\
Gemini 3.1 Pro & 530 & 833 & 474 & 1{,}837 & 3.06 & 17{,}941 & 14.75\% \\
Muse Spark 1.2 & 446 & 595 & 325 & 1{,}366 & 2.28 & 16{,}049 & 8.75\% \\
Claude Opus 5 & 549 & 892 & 770 & 2{,}211 & 3.69 & 18{,}225 & 16.38\% \\
\bottomrule
\end{tabular}
\caption{Hallucination axis: severity-tiered fabrication event counts, per-run rate, the field-level
assertion total underlying the flat rate, and the superseded flat rate itself. The flat rate is computed
from a separate field-level false-positive tally, not from the Total column, so it cannot be recovered as
Total divided by Asserted.}
\label{tab:hallucination}
\end{table}

No model comes close to zero on the tier that matters most clinically. Across all four models, most
fabrications in this tier are knowledge failures on the inferred generic name, the one field never
printed on the sheet, rather than an invented drug read directly off the page: in the per-field
fabrication logs underlying this tier, an invented active ingredient on an otherwise-real row is the
more common event, more so than a fully invented drug row, though the table reports only their combined
T1 total rather than this split. Because the accepted alternatives for each ingredient are now declared
by the ground truth row itself (Section~\ref{sec:medicine-cell-grading}), a correct ingredient named
under an accepted alternative is no longer charged here, and the residue in this tier is closer to a
genuine knowledge failure than the corresponding count in any earlier iteration of this benchmark.

Muse Spark 1.2 records the fewest fabrications in every column, including the flat rate, though its
margin over GPT-5.6 on the flat rate is under two tenths of a point. The two views of the medication
tier and of overall volume do not agree: Gemini 3.1 Pro fabricates fewer medication-tier entries than
GPT-5.6 yet more fabrications in total and at a substantially higher flat rate, because it asserts more
fields overall. Claude Opus 5 is the least restrained on every measure, and its non-clinical
tier alone exceeds the next-highest model's by more than sixty per cent.

\subsection{Engagement}

\begin{table}[t]
\centering
\footnotesize
\setlength{\tabcolsep}{3pt}
\begin{tabular}{@{}l r r r r r r@{}}
\toprule
& \multicolumn{2}{c}{Hard present ($n$=1{,}740)} & \multicolumn{2}{c}{Clear present ($n$=19{,}665)} & \multicolumn{2}{c}{Absent ($n\approx$11{,}215)} \\
\cmidrule(lr){2-3} \cmidrule(lr){4-5} \cmidrule(lr){6-7}
Model & Gave up & Attempt & Dropped & Attempt & Filled & Silent \\
\midrule
GPT-5.6 & 386 & 77.82\% & 2{,}227 & 88.68\% & 833 & 92.57\% \\
Gemini 3.1 Pro & 337 & 80.63\% & 2{,}330 & 88.15\% & 1{,}920 & 82.88\% \\
Muse Spark 1.2 & 421 & 75.80\% & 2{,}784 & 85.84\% & 971 & 91.34\% \\
Claude Opus 5 & 346 & 80.11\% & 2{,}069 & 89.48\% & 2{,}316 & 79.35\% \\
\bottomrule
\end{tabular}
\caption{Engagement axis: attempt rates on recoverable hard fields and on clear fields, and the silence
rate on fields the sheet leaves blank. The hard and clear populations (in parentheses) are fixed by
ground truth and identical across every model. The absent population is not: it is enumerated over the
union of the fields ground truth holds and the fields the model asserted, so a model that reaches for
more fields is measured against marginally more blanks. It varies from 11,213 to 11,217 across the four
models, a spread too small to affect the rates shown but stated here rather than implied away.}
\label{tab:engagement}
\end{table}

Every model attempts clear fields more often than hard ones; since every hard field carries a value a
human recovered, each such blank is a miss rather than caution. Gemini 3.1 Pro posts the highest
hard-field attempt rate, but its margin over Claude Opus 5 is half a point, so the two are better read as
a pair at the top than as first and second. GPT-5.6 sits some three points behind them and Muse Spark 1.2
a further two behind that.

On clear fields the ranking reshuffles and tightens: Claude Opus 5 attempts most often, GPT-5.6 and
Gemini 3.1 Pro follow within a point and a half of it, and only Muse Spark 1.2 separates from the group.
The gap between a model's clear- and hard-field attempt rates is what distinguishes them: it is widest
for GPT-5.6 and narrowest for Gemini 3.1 Pro, at roughly eleven points against seven and a half, so
Gemini's willingness carries over to degraded handwriting more consistently than GPT-5.6's does, even
though GPT-5.6 is the better reader of clear content by correctness.

Where silence is the correct response the ordering inverts almost completely. GPT-5.6 stays quiet most
often, with Muse Spark 1.2 a little over a point behind; the two models most willing to attempt hard
fields are also the two least willing to leave a blank alone, Claude Opus 5 filling nearly three times
as many blank fields as GPT-5.6 and Gemini 3.1 Pro more than twice as many. Engagement and restraint pull
against one another here, and no model holds both ends.

\subsection{Robustness}

\begin{table}[t]
\centering
\footnotesize
\setlength{\tabcolsep}{3pt}
\begin{tabular}{@{}l r r r r r r r@{}}
\toprule
& \multicolumn{3}{c}{Delivered} & \multicolumn{3}{c}{Attempted only} & \\
\cmidrule(lr){2-4} \cmidrule(lr){5-7}
Model & Hard & Clear & Drop & Hard & Clear & Drop & Parity \\
\midrule
GPT-5.6 & 49.23\% & 69.50\% & $+$20.27\% & 63.26\% & 78.38\% & $+$15.11\% & 89.14\% \\
Gemini 3.1 Pro & 51.03\% & 69.48\% & $+$18.45\% & 63.28\% & 78.82\% & $+$15.53\% & 92.48\% \\
Muse Spark 1.2 & 48.08\% & 64.03\% & $+$15.96\% & 63.42\% & 74.60\% & $+$11.17\% & 89.96\% \\
Claude Opus 5 & 46.35\% & 65.18\% & $+$18.82\% & 57.86\% & 72.84\% & $+$14.98\% & 90.64\% \\
\bottomrule
\end{tabular}
\caption{Robustness axis: hard- and clear-field performance and the drop between them, each in a
delivered form (skips charged as zeros) and an attempted-only form, with engagement parity as context.
The hard-field population ($n$=1{,}740) is fixed by ground truth and identical across every model; the
attempted-only columns are computed over each model's own attempted subset, which is why they are not
comparable across models without the parity column beside them.}
\label{tab:robustness}
\end{table}

The delivered and attempted-only rankings disagree, and the attempted-only view is close to
uninformative on its own. Under the delivered form, Gemini 3.1 Pro posts the highest hard-field score,
ahead of GPT-5.6 and Muse Spark 1.2 in that order, with Claude Opus 5 lowest. Read attempted-only, three
of the four models fall within a sixth of a point of one another and the nominal lead passes to Muse
Spark 1.2, the model with the lowest hard-field attempt rate in Engagement.

That inversion is a selection artefact rather than a finding: a model that declines the hard fields it
doubts most is scored only on the ones it kept, so its attempted-only figure is measured over an easier
subset. Once every skip is charged as a zero, Muse Spark 1.2 falls to third, though it retains the
smallest delivered drop of the four because its clear-field score is also the weakest, which narrows the
gap from above rather than lifting it from below. Gemini 3.1 Pro's lead is the more durable, since it
also holds the highest engagement parity, so little of it can be attributed to selective skipping.
GPT-5.6 shows the largest delivered drop and the lowest parity of the four, so its hard-field standing
depends more than any other model's on which fields it chose to leave blank. Claude Opus 5 is lowest on
delivered hard fields and lowest on both attempted-only columns by a clear margin, and is the weakest
reader of difficult handwriting under either accounting.

\subsection{Stability and Efficiency Modifiers}

\begin{table}[t]
\centering
\small
\begin{tabular}{@{}l r r r r r@{}}
\toprule
Model & Correctness mean & Std & Min & Max & Spread \\
\midrule
GPT-5.6 & 68.10\% & 14.49\% & 23.18\% & 98.31\% & 75.13\% \\
Gemini 3.1 Pro & 66.97\% & 15.51\% & 13.04\% & 96.97\% & 83.93\% \\
Muse Spark 1.2 & 64.45\% & 14.75\% & 0.00\% & 93.93\% & 93.93\% \\
Claude Opus 5 & 61.67\% & 14.58\% & 0.00\% & 94.52\% & 94.52\% \\
\bottomrule
\end{tabular}
\caption{Stability modifier: run-to-run spread of the primary correctness metric over 600 cold runs per
model.}
\label{tab:stability}
\end{table}

\begin{table}[t]
\centering
\small
\begin{tabular}{@{}l r r r r r@{}}
\toprule
Model & Time/run & Cost/run & Total cost & Tokens in/out & Parse errors \\
\midrule
Claude Opus 5 & 40.9 s & \$0.1233 & \$73.97 & 9{,}235 / 3{,}085 & 0.50\% \\
Gemini 3.1 Pro & 52.9 s & \$0.0696 & \$41.76 & 6{,}249 / 4{,}888 & 0\% \\
Muse Spark 1.2 & 109.3 s & \$0.0401 & \$24.07 & 6{,}787 / 7{,}588 & 0.50\% \\
GPT-5.6 & 153.4 s & \$0.2416 & \$144.97 & 9{,}257 / 7{,}090 & 0\% \\
\bottomrule
\end{tabular}
\caption{Efficiency modifier: latency, cost, tokens, and parse-error rate per run.}
\label{tab:efficiency}
\end{table}

Neither modifier lines up with the quality ordering. GPT-5.6 is the most accurate model in this pool and
also the steadiest, though three of the four models sit within three tenths of a point of one another on
standard deviation, so only Gemini 3.1 Pro separates: it is second on average correctness and the least
steady of the four, with the widest spread of any model that never scored zero. Both Claude Opus 5 and
Muse Spark 1.2 hit a complete zero on at least one run, which fixes their spread at their maximum and
makes that column incomparable with the other two models'.

Muse Spark 1.2 is the cheapest model per run and Claude Opus 5 the fastest, while GPT-5.6 is the slowest
and by far the most expensive, at roughly six times Muse Spark 1.2's cost per run. That expense is driven
by a higher per-token price rather than by writing more output, since Muse Spark 1.2 emits marginally more
output tokens per run than GPT-5.6 does. Claude Opus 5 and Muse Spark 1.2 share the only parse failures in
this pool: for each of these two models, all three of its independently scheduled cold runs on one
image, the largest in the set, failed to parse. These are three separate runs, not three retries of one
run; a parse failure is never retried (Section~3), so each counts as its own independent failure rather
than a compounding retry sequence. Each such run was still scored zero under the protocol in Section~3
rather than excluded; no other model produced an unparseable response.

\subsection{Cross-Axis Synthesis}

Reading the four axes together, Gemini 3.1 Pro leads two of the four, engagement and robustness; GPT-5.6
leads correctness and Muse Spark 1.2 leads hallucination; Claude Opus 5 leads none.

\begin{itemize}
  \item GPT-5.6 leads overall and medicine-only correctness, is the steadiest from run to run, and stays
  silent on blank fields more often than any other model. It trails on engagement and robustness, with
  the widest hard-to-clear attempt-rate gap and the lowest engagement parity of the four, so its
  hard-field standing rests more than any other model's on which fields it declined. It is by far the
  slowest and most expensive model to run.
  \item Gemini 3.1 Pro leads both engagement and robustness: the highest hard-field attempt rate and the
  highest delivered hard-field score, backed by the highest engagement parity, so neither lead is an
  artefact of selective skipping, and it holds the narrowest gap between clear- and hard-field
  willingness. It also nominally leads the pooled clinical-plus-non-clinical correctness split, by a
  margin too small to separate it from Muse Spark 1.2. It is the least stable model from run to run and
  the second-least restrained on blank fields.
  \item Muse Spark 1.2 fabricates least on every hallucination measure, including the flat rate, but
  posts the lowest hard-field attempt rate and the lowest clear-field attempt rate of the four. Its
  attempted-only hard-field lead, the only measure on which it leads besides hallucination, is a
  selection effect of having skipped the fields it doubted most, and its delivered hard-field score is
  third. It is the cheapest model to run.
  \item Claude Opus 5 trails on correctness across every split, posts the lowest delivered hard-field
  score and the lowest attempted-only figures on both hard and clear fields, and is the least restrained
  model on blank fields. It leads the clear-field attempt rate and is within half a point of the top on
  hard fields, so it is willing without being accurate. It is the fastest model to run.
\end{itemize}

This fragmentation, rather than any leaderboard position, is the finding: reading, avoiding invention,
engaging with hard content, and degrading gracefully are separable skills that no model masters
uniformly, even when one model, Gemini 3.1 Pro, leads two of the four axes. The axes also trade off
against one another rather than merely differing: the two models most willing to attempt hard fields are
the two least willing to leave a blank alone, and the model that fabricates least is the least willing to
attempt anything. Several margins are narrow enough that they should not be read as separations at all,
and one axis ranking inverts entirely depending on whether declined fields are charged as zeros. This
evaluation should not be read as a settled ranking.

\section{Reproducibility}

Every number in Section~6 is produced without re-running any model: the benchmark computation performs a
deterministic regrouping of stored per-field scores into the four axes, and a companion test suite
verifies that this regrouping reproduces those stored scores bit-for-bit. The scoring rules in Sections~4
and 5 have been checked field by field against the stored evaluation artifact.

Scoring itself involves no model in the loop, so re-scoring the same predictions against the same ground
truth reproduces every per-record value exactly; this was confirmed by re-running one batch in full and
comparing each record's axis values and presence counts against the first pass. Each stored record
carries the version of the scoring rules that produced it, so a number can always be traced to the rule
set behind it, and records produced under different versions are never pooled. Every figure in Section~6
comes from a single re-score of the whole corpus under one version of the rules.

\section{Conclusion}

RxScribe Bench evaluates four separable skills a prescription-reading model must get right: correctness,
restraint from invention, engagement with legible content, and graceful degradation on hard handwriting.
Across four frontier models over 200 real prescriptions with three cold runs each, no model approaches
production readiness. All four read the medicine table considerably worse than the rest of the sheet, by
twenty to twenty-nine points, and every model fabricates within every severity tier, with the most
dangerous tier's fabrications attributable mostly to gaps in pharmacological knowledge rather than to
misreading. All four also decline more often on recoverable difficult handwriting than on clear
handwriting, which renders those blanks missed content rather than caution. No single model leads
throughout: the most accurate reader, GPT-5.6, is also the slowest and most expensive to run; the model
most willing to attempt hard fields, Gemini 3.1 Pro, is the least stable from run to run; the fastest
model to run, Claude Opus 5, is the weakest reader on every correctness split and the least restrained on
fields the page leaves blank; and the model that fabricates least, Muse Spark 1.2, is the least willing to
attempt anything and owes its attempted-only robustness score to having already skipped the hard fields
it doubted most. This shape, not any single ranking, is the benchmark's principal result, and should not
be read as more certain than 200 prescriptions can support.

\section*{Declaration on the Use of AI Tools}

The authors employed large language models at three points in producing this work, namely editing prose
for grammar and clarity, assisting with data analysis, and drafting statistical summaries. None of these
tools had access to the underlying evaluation pipeline or the artifacts it produced. Every number,
claim, and conclusion presented herein was independently verified by the authors against those
artifacts, and the
authors bear sole responsibility for what is written.

\end{document}